\documentclass{INTERSPEECH2023}

\interspeechcameraready

\usepackage{mathtools}
\usepackage{multicol}
\usepackage[table,xcdraw]{xcolor}
\usepackage{amsmath,graphicx}
\usepackage{adjustbox,multirow}
\usepackage{hyperref}
\usepackage[symbol]{footmisc}

\usepackage[table]{xcolor}
\usepackage{soul}

\newcommand{\expresso}{\textsc{Expresso}}

\title{\textsc{Expresso}: A Benchmark and Analysis of \\ Discrete Expressive Speech Resynthesis}

\name{Tu Anh Nguyen$^{*,1}$\thanks{$^{*+}$Core contribution as first and last authors.}, Wei-Ning Hsu$^{*,1}$, Antony D'Avirro$^{*,1}$, Bowen Shi$^{*,1}$, Itai Gat$^1$, \\
Maryam Fazel-Zarani$^1$, Tal Remez$^1$, Jade Copet$^1$, Gabriel Synnaeve$^1$, \\ Michael Hassid$^{1,2}$, Felix Kreuk$^1$,
Yossi Adi$^{+,1,2}$, Emmanuel Dupoux$^{+,1}$}

\address{$^1$Meta AI\\ 
$^2$The Hebrew University of Jerusalem}

\email{\{ntuanh,adiyoss,dpx\}@meta.com}

\begin{document}

\maketitle

\begin{abstract}
Recent work has shown that it is possible to resynthesize high-quality speech based, not on text, but on low bitrate discrete units that have been learned in a self-supervised fashion and can therefore capture expressive aspects of speech that are hard to transcribe (prosody, voice styles, non-verbal vocalization). 
The adoption of these methods is still limited by the fact that most speech synthesis datasets are read, severely limiting spontaneity and expressivity. Here, we introduce \expresso, a high-quality expressive speech dataset for textless speech synthesis that includes both read speech and improvised dialogues rendered in 26 spontaneous expressive styles. We illustrate the challenges and potentials of this dataset with an \textit{expressive resynthesis benchmark} where the task is to encode the input in low-bitrate units and resynthesize it in a target voice while preserving content and style. We evaluate resynthesis quality with automatic metrics for different self-supervised discrete encoders, and explore tradeoffs between quality, bitrate and invariance to speaker and style. All the dataset, evaluation metrics and baseline models are open source\footnote{\url{https://speechbot.github.io/expresso/}}.

\end{abstract}

\vspace*{-3mm}

\section{Introduction and related work}
\label{sec:intro}
Speech synthesis has been traditionally approached as a mapping between text and speech. This has a series of limiting consequences: text is an impoverished representation of language, that does not specify many expressive dimensions: rhythm, intonation, emotion, emphasis, and often fails to encode non-verbals like laughter, cries, lip smacks, etc. As a result, speech synthesizers typically resort to standard read speech as the main target, which severely limits the expressivity of AI systems.

Everything changed with the advent of Self Supervised Learning (SSL) speech models \cite{hsu21hubert, chung21w2vbert}, which enabled to build discrete representations for speech without needing any textual annotation \cite{lakhotia21gslm, borsos22audiolm, gat2023augmentation}. Because such representations can be learned from much more diverse audio than read speech (conversational, casual speech), this opens up the possibility to build more expressive systems based on SSL units instead of text.
Recently, \cite{wang2023valle, kharitonov2023speartts} utilize neural audio codecs \cite{defossez2022highfi, zeghidour2021soundstream} to encode speech features into codes and generate natural speech from textual input. However, these models still rely on large-scale read speech that lacks expressivity.
One of the roadblocks in 
building such expressive systems
is the lack of datasets that are sufficiently expressive and of high audio quality for learning a synthesizer. Most existing expressive datasets (e.g., EmoV \cite{adigwe2018emov}) have used expressive reading, where voice actors read a (neutral) sentence in different expressions (happy, sad, etc.). This method puts the voice actors in an artificial situation, resulting in not very plausible rendering of these expressions. 
In this work, we also reproduced this protocol to create an expressive speech dataset but added a section based on conversational improvisation. 
The two actors are prompted with a situation and a character (e.g., two drivers involved in a car accident, a parent and a child, etc.) and improvise a dialogue impersonating their characters. This yields much more realistic and casual speech, with spontaneous hesitations, laughter, etc., that would be extremely hard to transcribe accurately, but, which can be in principle captured by SSL units. 

Next, we illustrate the potential of this dataset by setting up the task of discrete expressive resynthesis. As in discrete resynthesis \cite{polyak21resynt, maimon2022speaking}, it consists in taking audio as input, encoding it in low bitrate discrete units, and synthesizing the same content with a different target voice.
In this work, we additionally include the task of preserving expressive style.
We introduce automatic metrics for content and style preservation and evaluate a HuBERT~\cite{hsu21hubert} encoder trained on a masked prediction objective followed by k-means clustering, which we compare to Encodec~\cite{defossez2022highfi}, a compression model which acts as a high bitrate baseline. We compare different pretraining sources for the HuBERT units based on public datasets of read speech and/or spontaneous speech. Synthesis is done with a units-based HiFi-GAN vocoder~\cite{polyak21resynt} conditioned on speaker or speaker and style.

\vspace{-0.2cm}
\section{The \expresso~ dataset}
The \expresso~dataset consists of 47 hours of expressive speech from 4 speakers of North American English. The dataset is divided into two main sections: an \textit{expressive reading} section (37\% of the corpus) where actors read short prompts in a parallel fashion in 7 different styles, with additional long-form and emphasis material (see \ref{sec:reading}), and an \textit{improvised dialog} section (72\% of the corpus) where pairs of actors are prompted to improvise a conversation in a fictive setting that illustrates one of 25 specified styles (see \ref{sec:improv}). In a small additional singing section, actors sing a few of their favorite songs. 

The 26 different styles were chosen for their universality/recognizability (i.e. common emotions like happiness, sadness), utility for current/anticipated speech applications (i.e. whispered, enunciated speech), and to elicit the large range of possible vocalizations of the human voice (including addressing or imitating a child or an animal, and non-verbals like grunting, coughing, whistling, etc., see Table \ref{tab:expresso} for a full list).

Data was recorded in a professional recording studio with minimal background noise at 48kHz/24bit. The files for read speech and singing are in a mono wav format; and for the dialog section in stereo (one channel per actor), where the original flow of turn-taking is preserved.

\begin{table}[t!] \footnotesize
\caption{\expresso's expressive styles. * singing is the only improvised style that is not in dialogue format.}
\vspace{-1.7em}
\center{
\begin{tabular}{lp{1cm}p{1.3cm}c}
\toprule
Style  &Read (min) & Improvised (min) & total hours \\
\midrule
angry &- &82 &1.4\\ 
animal &- &27 &0.4\\
animal\_directed &- &32 &0.5\\
awe &- &92 &1.5\\
bored &- & 92 &1.5\\
calm &- &93 &1.6\\
child &- &28 &0.4\\
child\_directed &- &38 &0.6\\
confused &94 &66 &2.7\\
default &133 &158 &4.9\\
desire &- &92 &1.5\\
disgusted &- &118 &2.0\\
enunciated &116 &62 &3.0\\
fast &- &98 &1.6\\
fearful &- &98 &1.6\\
happy &74 &92 &2.8\\
laughing &94 &103 &3.3\\
narration &21 &76 &1.6\\
non\_verbal &- &32 &0.5\\
projected &- &94 &1.6\\
sad &81 &101 &3.0\\
sarcastic &- &106 &1.8\\
singing* &- &4 &.07\\
sleepy &- &93 &1.5\\
sympathetic &- &100 &1.7\\
whisper &79 &86 &2.8\\
\bottomrule
\end{tabular}
}\label{tab:expresso}
\vspace{-2.0em}
\end{table}

\vspace{-0.5em}
\subsection{Expressive reading}\label{sec:reading}
\vspace{-0.5em}
Seven of the styles (confused, default, enunciated, happy, laughing, sad, whisper) were applied in a parallel fashion to the same set of prompts, so content did not necessarily reflect the emotion being conveyed, and we relied on the actor’s expertise to convey the desired style. Written instructions were delivered for each style describing the upper and lower performative bounds of the style, some including video examples. 

Aside from a small corpus of shared essential lines (greetings, common phrases, numbers, letters), each speaker had a unique script in order to maximize linguistic diversity across the entire dataset. In total, the written corpus contains roughly 21,000 words over 2,400 unique lines. Material was scraped from open-source datasets like Wikipedia and commissioned datasets containing voice-assistant style utterances and then proofread and scrubbed for PII. Although not specifically balanced for phonetic coverage, the corpus was tuned both overall and per-speaker for a desired ratio of statements, questions, exclamations, jokes, etc.

\noindent\textbf{Contrastive emphasis.} We enclose certain words/spans in asterisks to denote emphasis, designed to convey contrastive focus in the reading of an utterance. Actors were trained to read this syntax with the desired prosodic effect. These occur in isolation throughout the read-speech corpus, but also in a subset of each speaker's lines labeled ``emphasis'' where the same line is repeated 2 to 4 times with contrastive emphasis placed on different words/spans.

\noindent\textbf{Long-form material.} To capture longer-range prosodic dependencies, each speaker’s script contains one news article (read in the ``default'' style) and one long-form narrative piece (read in the ``narration'' style), roughly totalling 100 lines per speaker. 

\begin{table}[t!]
\caption{Encoder and Tokenizer used for our discrete units. Bitrate is log$_2$(codebook size) $\times$ n units per sec in BPS. Mean within- and across-speaker ABX discrimination scores, resp., on 1-hot vectors and units' centroids. PNMI is mutual information between units and phonemes. For Encodec RVQ8, we concatenate multiple codebooks for ABX. PNMI is not available for this tokenizer. 
\label{tab:units}}
\vspace{-1.1em}
\centering
\setlength{\tabcolsep}{2pt}
\resizebox{1.\linewidth}{!}{
\begin{tabular}{llccccccc}
\toprule
 \multirow{2}{*}{Model} & \multirow{2}{*}{Tokenizer} & \multirow{2}{*}{BPS} & \multicolumn{2}{c}{ABX 1-hot ($\downarrow$)} & \multicolumn{2}{c}{ABX-centr. ($\downarrow$)} & \multicolumn{2}{c}{PNMI ($\uparrow$)} \\
\cmidrule(lr{.75em}){4-5} \cmidrule(lr{.75em}){6-7} \cmidrule(lr{.75em}){8-9}
 &  &  & LS & Fisher & LS & Fisher & LS & Fisher \\
\midrule
\midrule
\multirow{2}{*}{\begin{tabular}[c]{@{}c@{}}HuBERT\\(LS960)\end{tabular}} & KM500 (LS960) & 450 & \textbf{8.98} & 15.45 & 5.28 & 10.84 & 67.57 & 48.42 \\
& KM2000 (Expr) & 550 & 11.00 & 17.81 & \textbf{4.32} & 9.07 & 73.23 & 56.33 \\
\midrule
\multirow{2}{*}{\begin{tabular}[c]{@{}c@{}}HuBERT\\(Mix1)\end{tabular}} & KM2000 (Mix1) & 550 & 10.27 & \textbf{14.96} & 4.92 & 8.93 & 72.36 & 55.77 \\
 & KM2000 (Expr) & 550 & 10.50 & 15.27 & 4.39 & \textbf{8.04} & \textbf{74.37} & \textbf{59.14} \\
\midrule
\multirow{2}{*}{Encodec} & RVQ1 & 500 & 45.47 & 44.29 & 26.10 & 29.57 & 21.42 & 13.17 \\
 & RVQ8 & 4000 & 41.38 & 40.50 & 20.20 & 25.60 & - & - \\         
\midrule
\end{tabular}}
\vspace{-1.5em}
\end{table}

\subsection{Improvised dialogs section}\label{sec:improv}
Dialogs were elicited via a set of situational prompts designed to evoke the desired styles or emotions. Some prompts resemble voice-application domains such as reporting the weather, navigation, information retrieval, while others are more open-ended scenarios; some of them were proposed by the actors.

Tradeoffs were made to capture usable data while preserving the feel of natural conversation, i.e. actors were recorded in separate booths but watching each other over video conference. A small number of dialogs ($<$10) were interrupted by the studio director to provide notes. These dialogs were edited to remove the pause and maintain a more natural flow.

\subsection{Singing section}
Each speaker recorded several versions of popular nursery rhymes and public domain songs. The original recording had more data (93 minutes total) but could not be shared because the songs turned out not to be in the public domain. 

\subsection{Data preparation}
The raw dataset contains mostly pre-cut segments of 3-4 seconds for the read section, except for long-form ones, and long waveforms ranging from 2 to 10 minutes for dialogs section. 
For the purpose of speech synthesis, we cut long files into segments of 15 seconds. We split the dataset into train/dev/test subsets such that each speaker-style contains roughly 60s in dev/test splits, resulting in 1.5 hours of speech in each subset.

\begin{table*}[t!]\footnotesize
\centering
\caption{Content preservation evaluation: WERs (\%) of speech resynthesized by our models. We bold the best HuBERT or best Encodec model within each column. We denote speaker conditioning as S, and speaker with expression conditioning as S\_E. Results are reported for Expresso (E), VCTK (V) and Fisher (Fish). \label{tab:wers}}
\vspace{-.7em}
\resizebox{\linewidth}{!}{
\begin{tabular}{llll|ccc|cc|cccccc|cc}
\toprule
 &  &  &  & \multicolumn{5}{c|}{In Domain Source} & \multicolumn{8}{c}{Out of Domain Source} \\
 \hline
 &  &  &  & \multicolumn{3}{c}{Same Speaker} & \multicolumn{2}{c|}{Swapped Speaker} &  &  &  &  &  &  &  &  \\
 \cline{5-17} 
 &  &  & Src. & LJ & V & E & V & E & LS & Fish & LS & Fish & LS & Fish & LS & Fish \\
 \cline{5-17} 
 &  &  & Tgt. & LJ & V & E & V & E & LJ & LJ & V & V & E & E & \multicolumn{2}{c}{Orig Voices} \\
\hline 
Model & Tokenizer & Data & Cond. & \multicolumn{13}{c}{WER} \\ \hline
\textit{Original audio} & \_ & \_ & \_ & \textit{2.04} & \textit{1.74} & \textit{14.76} & \_ & \_ & \_ & \_ & \_ & \_ & \_ & \_ & \textit{3.55} & \textit{30.26} \\ 
\hline
\multicolumn{1}{c|}{\multirow{6}{*}{\begin{tabular}[c]{@{}c@{}} HuBERT\\ (LS960) \end{tabular}}} & \multirow{3}{*}{\begin{tabular}[c]{@{}c@{}} KM500\\(LS960)\end{tabular}} & \begin{tabular}[c]{@{}c@{}}LJ+V \end{tabular} & S & 3.12 & 6.85 & \_ & 7.20 & \_ & 11.56 & 50.13 & 10.83 & 48.26 & \_ & \_ & \_ & \_ \\ 
\multicolumn{1}{c|}{} &  & \begin{tabular}[c]{@{}c@{}}E+LJ+V\end{tabular} & S & 3.22 & 7.19 & 24.21 & 6.80 & 24.34 & 11.61 & 49.24 & 10.42 & 46.63 & 10.93 & 47.18 & \_ & \_ \\ 
\multicolumn{1}{c|}{} &  & \begin{tabular}[c]{@{}c@{}}E+LJ+V\end{tabular} & \begin{tabular}[c]{@{}c@{}} S\_E \end{tabular} & 3.65 & 6.76 & 23.65 & 7.79 & 24.82 & 12.00 & 50.49 & 10.75 & 47.72 & 10.57 & 46.57 & \_ & \_ \\ \cline{2-17} 
\multicolumn{1}{c|}{} & \multirow{3}{*}{\begin{tabular}[c]{@{}c@{}} KM2000 (E) \end{tabular}} & LJ+V & S & 2.98 & 7.15 & \_ & 7.09 & \_ & 10.95 & 47.44 & 9.98 & 47.06 & \_ & \_ & \_ & \_ \\
\multicolumn{1}{c|}{} &  & \begin{tabular}[c]{@{}c@{}}E+LJ+V\end{tabular} & S & 3.41 & 7.09 & 21.64 & 7.01 & 22.80 & 10.80 & 46.90 & 10.20 & 45.79 & 10.61 & 46.77 & \_ & \_ \\ 
\multicolumn{1}{c|}{} &  & \begin{tabular}[c]{@{}c@{}}E+LJ+V\end{tabular} & \begin{tabular}[c]{@{}c@{}} S\_E \end{tabular} & 2.83 & 6.48 & 22.35 & 6.56 & 21.92 & 10.34 & 45.93 & 9.72 & 45.08 & 9.52 & 43.13 & \_ & \_ \\ 
\hline
\multicolumn{1}{c|}{\multirow{6}{*}{\begin{tabular}[c]{@{}c@{}}HuBERT\\(Mix1)\end{tabular}}} & \multirow{3}{*}{\begin{tabular}[c]{@{}c@{}}KM2000 (Mix1)\end{tabular}} & LJ+V & S & \textbf{2.60} & 6.98 & \_ & 7.60 & \_ & 9.60 & 41.78 & 8.34 & 40.53 & \_ & \_ & \_ & \_ \\ 
\multicolumn{1}{c|}{} &  & \begin{tabular}[c]{@{}c@{}}E+LJ+V\end{tabular} & S & 2.80 & 6.84 & 21.25 & 7.20 & 22.52 & 9.17 & 40.61 & 8.38 & 38.91 & 9.76 & 42.87 & \_ & \_ \\ 
\multicolumn{1}{c|}{} &  & \begin{tabular}[c]{@{}c@{}}E+LJ+V\end{tabular} & \begin{tabular}[c]{@{}c@{}} S\_E \end{tabular} & 2.85 & 7.17 & 20.36 & 7.33 & 20.81 & 9.50 & 41.09 & 8.92 & 40.82 & 8.39 & 38.47 & \_ & \_ \\ \cline{2-17} 
\multicolumn{1}{c|}{} & \multirow{3}{*}{\begin{tabular}[c]{@{}c@{}}KM2000 (E)\end{tabular}} & LJ+VCTK & S & 2.77 & 5.60 & \_ & 5.89 & \_ & 9.48 & 41.42 & 8.39 & 40.81 & \_ & \_ & \_ & \_ \\
\multicolumn{1}{c|}{} &  & \begin{tabular}[c]{@{}c@{}}E+LJ+V\end{tabular} & S & 2.95 & \textbf{4.85} & 20.64 & \textbf{5.07} & 21.01 & \textbf{9.04} & 39.62 & 7.91 & 38.45 & 8.46 & 39.84 & \_ & \_ \\
\multicolumn{1}{c|}{} &  & \begin{tabular}[c]{@{}c@{}}E+LJ+V\end{tabular} & \begin{tabular}[c]{@{}c@{}} S\_E \end{tabular} & 3.05 & 5.48 & \textbf{19.52} & 5.59 & \textbf{20.27} & 9.20 & \textbf{38.79} & \textbf{7.75} & \textbf{37.48} & \textbf{8.00} & \textbf{36.67} & \_ & \_ \\ \hline
\multicolumn{2}{l}{Encodec (RVQ-1)} & \_ & None & 5.52 & 17.46 & 34.36 & \_ & \_ & \_ & \_ & \_ & \_ & \_ & \_ & 18.88 & 60.68 \\
\multicolumn{2}{l}{Encodec (RVQ-8)} & \_ & None & \textbf{2.20} & \textbf{2.52} & \textbf{16.85} & \_ & \_ & \_ & \_ & \_ & \_ & \_ & \_ & \textbf{4.62} & \textbf{35.64} \\ 
\bottomrule
\end{tabular}}
\vspace{-1.8em}
\end{table*}

\section{Method}
\noindent {\bf Additional datasets.} We refer to the read section of \expresso~as Exp-R and the improvised section as Exp-I. To train the units and vocoder, and to evaluate the results, we use additional open source datasets: LJspeech \cite{ljspeech17} (LJ), VCTK \cite{Veaux2017CSTRVCTK}, Librispeech dev-other \cite{librispeech} (LS), Fisher \cite{Cieri2004TheFC} and EmoV-DB (EmoV) \cite{adigwe2018emov}.

\noindent {\bf Evaluation metrics}
For evaluating the discrete units, we compute their bitrate, ABX discrimination (the probability that the DTW distance between minimally different triphones like /bit/ versus /bet/ are more distant to one another than two instances of the same triphone \cite{abxschatz}) both on the 1-hot representations (as in \cite{nguyen2020zero}), and using the dense embedding corresponding to the centroid of the units, and PNMI (phone-normalized mutual information between units $z$ and ground truth phonemes $y$: $I(y;z)/H(y)$ as in \cite{hsu21hubert}).

For evaluating the quality of resynthesized speech, we build automatic metrics for the preservation of content, pitch, and expressivity. As in \cite{polyak21resynt}, content preservation is evaluated by running a publicly available Automatic Speech Recognition (ASR) model \cite{xu2021self} on the resynthesized sentence and computing the Word Error Rate (WER) relative to the transcription of the input sentence\footnote{\url{https://dl.fbaipublicfiles.com/fairseq/wav2vec/wav2vec_vox_960h_pl.pt}}. We run this on in-domain inputs (LJ, VCTK and Exp-R) and out-of-domain inputs (LS and Fisher). Pitch preservation is evaluated by computing F0 Frame Error (FFE), which measures the percentage of frames with a deviation of more than 20\% in pitch value between the input and resynthesized output. Expressivity preservation is computed by training an expressive style classifier\footnote{We fine-tune the wav2vec2 base model \cite{baevski20w2v2} on a 26 style classes audio classification task (as in the SUPERB benchmark \cite{Yang2021SUPERBSP}) using Huggingface transformers library \cite{wolf-etal-2020-transformers}).} on the train set of \expresso~and applying it to resynthesized versions of its dev set. These classifiers are also run on the original data for comparison.

\vspace{-0.2cm}
\section{Models}
\vspace{-0.2cm}
\subsection{Unit encoding}
For unit encoding, we compared three models: two HuBERT-based, one Encodec-based. The HuBERT models use the same architecture (HuBERT base with 12 Tranformer layers), but are trained on different corpora. HuBERT-LS960 was trained on LibriSpeech 960 as in \cite{lakhotia21gslm}. We used the available model in textless-lib \cite{Kharitonov2022textlesslib}. We use HuBERT-Mix1 from \cite{hsu2022revise}, which was trained with a more varied mixture of datasets: an 8 language subset of VoxPopuli \cite{wang-etal-2021-voxpopuli} (167K h), Common Voice \cite{ardila-etal-2020-commonvoice} (4K h) and Multilingual LibriSpeech (MLS) \cite{Pratap2020MLSAL} (50K h), totalling 221K hours. For quantization, we trained k-means models on HuBERT features either on a subset of HuBERT pre-training dataset or on \expresso, with k=500 on LS960 or k=2000 on other datasets. 
The Encodec model is from \cite{defossez2022highfi}, we used two models with 1 and 8 codebooks of cardinality 1024, which were trained on VoxPopuli400k English, People's Speech \cite{galvez21pplspeech}, LibriSpeech 960, LibriLight \cite{librilight} and Spotify \cite{clifton-etal-2020-spotify}. Training hyperparameters were identical to \cite{defossez2022highfi} except for having no audio normalization and using zero padding instead of reflect.

\vspace{-0.2cm}
\subsection{Vocoder}
For HuBERT units, we produce the waveform using HiFiGAN, which we train on the units presented above. For each set of units, we train either on LJ and VCTK, or on LJ, VCTK and \expresso. The vocoder is either conditioned on speaker ID (using a look-up table), or on speaker ID and expression ID (also using a look-up table). We distinguish the read and improvised versions of the expressions, yielding a total of 34 expressions. For Encodec units, we used the Encodec decoder to directly produce the waveform, and compared systems with 1 and 8 codebooks.



\begin{table*}[t!]\footnotesize
\centering
\caption{Expressive style classification accuracy using a pre-trained emotion classification model. We denote speaker conditioning as S, and speaker with expression conditioning as S\_E. Results are reported for Expresso (E) and VCTK (V).\label{tab:EMO}}
\vspace{-.7em}
\resizebox{0.9\linewidth}{!}{
\begin{tabular}{llll|cc|cc|cccc|c}
\toprule
\multicolumn{4}{c}{} & \multicolumn{2}{c}{Same} & \multicolumn{2}{c}{Swapped} & \multicolumn{4}{c}{Zero-shot} & Out of dom. \\
\hline
 &  &  & Src. & E\_R & E\_I & E\_R & E\_I & E\_R & E\_I & E\_R & E\_I & EmoV \\
 \cline{5-13} 
 &  &  & Tgt. & \multicolumn{1}{c}{E} & E & E & E & LJ & LJ & V & V & E \\ \hline
Model & Tokenizer & Data & Cond. & \multicolumn{9}{c}{Accuracy} \\ \hline
\multicolumn{1}{c}{\textit{Original Audio}} & \multicolumn{1}{c}{\_} & \_ & \_ & \textit{92.47} & \textit{75.69} & \_ & \_ & \_ & \_ & \_ & \_ & \textit{27.46} \\ \hline
\multicolumn{1}{c|}{\multirow{6}{*}{\begin{tabular}[c]{@{}c@{}}HuBERT \\ (LS960) \end{tabular}}} & \multicolumn{1}{c|}{\multirow{3}{*}{\begin{tabular}[c]{@{}c@{}}KM500\\(LS960) \end{tabular}}} & LJ+V & S & - & - & \_ & \_ & 13.36 & 9.14 & 26.42 & 12.43 &  \\
\multicolumn{1}{c|}{} & \multicolumn{1}{c|}{} & \begin{tabular}[c]{@{}c@{}}E+LJ+V\end{tabular} & S & 33.18 & 14.99 & 29.80 & 13.53 & 8.45 & 10.79 & 28.26 & 10.97 & 11.56 \\
\multicolumn{1}{c|}{} & \multicolumn{1}{c|}{} & \begin{tabular}[c]{@{}c@{}}E+LJ+V\end{tabular} & \begin{tabular}[c]{@{}c@{}}S\_E \end{tabular} & \textbf{81.57} & 58.68 & 81.26 & 56.12 & 23.96 & 36.93 & \textbf{56.07} & 28.15 & 43.35 \\ \cline{2-13} 
\multicolumn{1}{c|}{} & \multicolumn{1}{c|}{\multirow{3}{*}{\begin{tabular}[c]{@{}c@{}}KM2000 (E)\end{tabular}}} & LJ+V & S & - & - & \_ & \_ & 5.22 & 10.42 & 27.04 & 11.70 &  \\
\multicolumn{1}{c|}{} & \multicolumn{1}{c|}{} & \begin{tabular}[c]{@{}c@{}}E+LJ+V\end{tabular} & S & 39.17 & 23.95 & 33.95 & 19.38 & 11.67 & 12.07 & 27.34 & 14.63 & 9.25 \\
\multicolumn{1}{c|}{} & \multicolumn{1}{c|}{} & \begin{tabular}[c]{@{}c@{}}E+LJ+V\end{tabular} & \begin{tabular}[c]{@{}c@{}} S\_E \end{tabular} & 78.34 & \textbf{62.16} & 76.96 & 54.11 & 22.12 & 39.31 & 55.76 & \textbf{32.54} & 46.24 \\ \hline
\multicolumn{1}{c|}{\multirow{6}{*}{\begin{tabular}[c]{@{}c@{}}HuBERT\\ (Mix1)\end{tabular}}} & \multicolumn{1}{c|}{\multirow{3}{*}{\begin{tabular}[c]{@{}c@{}}KM2000 (Mix1)\end{tabular}}} & LJ+V & S & - & - & \_ & \_ & 7.99 & 9.32 & 27.50 & 11.52 &  \\
\multicolumn{1}{c|}{} & \multicolumn{1}{c|}{} & \begin{tabular}[c]{@{}c@{}}E+LJ+V\end{tabular} & S & 25.81 & 17.73 & 28.57 & 15.72 & 5.53 & 8.96 & 27.65 & 11.33 & 11.27 \\
\multicolumn{1}{c|}{} & \multicolumn{1}{c|}{} & \begin{tabular}[c]{@{}c@{}}E+LJ+V\end{tabular} & \begin{tabular}[c]{@{}c@{}} S\_E \end{tabular} & 78.80 & 61.06 & \textbf{81.41} & \textbf{58.14} & 31.64 & \textbf{40.77} & 40.86 & 31.63 & 48.27 \\ \cline{2-13} 
\multicolumn{1}{c|}{} & \multicolumn{1}{c|}{\multirow{3}{*}{\begin{tabular}[c]{@{}c@{}}KM2000 (E) \end{tabular}}} & LJ+V & S & - & - & \_ & \_ & 7.68 & 10.24 & 27.80 & 12.07 &  \\
\multicolumn{1}{c|}{} & \multicolumn{1}{c|}{} & \begin{tabular}[c]{@{}c@{}}E+LJ+V\end{tabular} & S & 37.02 & 16.82 & 35.33 & 16.09 & 5.99 & 9.69 & 26.73 & 11.52 & 14.45 \\
\multicolumn{1}{c|}{} & \multicolumn{1}{c|}{} & \begin{tabular}[c]{@{}c@{}}E+LJ+V\end{tabular} & \begin{tabular}[c]{@{}c@{}} S\_E\end{tabular} & 72.81 & \textbf{62.16} & 73.12 & 55.76 & \textbf{31.18} & 39.85 & 55.61 & 28.52 & \textbf{49.71} \\ \hline
\multicolumn{2}{l|}{Encodec (RVQ-1)} & \_ & None & 57.76 & 44.42 & \_ & \_ & \_ & \_ & \_ & \_ & 22.25 \\
\multicolumn{2}{l|}{Encodec (RVQ-8)} & \_ & None & \textbf{78.65} & \textbf{64.53} & \_ & \_ & \_ & \_ & \_ & \_ & \textbf{26.88} \\ 
\bottomrule
\end{tabular}}
\end{table*}

\begin{table*}[t!] \footnotesize
\centering
\caption{F0 Frame Error (FFE). Bold best absolute scores. We denote speaker conditioning as S, and speaker with expression conditioning as S\_E. Results are reported for Expresso (E), Expresso Read (E\_R), Expresso Improvised (E\_I), LJ, VCTK (V), and EMOV.\label{tab:FFE}}
\vspace{-.7em}
\resizebox{\linewidth}{!}{
\begin{tabular}{llll|cc|cc|cccc|c}
\toprule
\multicolumn{4}{c}{} & \multicolumn{2}{c}{Same} & \multicolumn{2}{c}{Swapped} & \multicolumn{4}{c}{Zero-shot} & OOD \\
\hline
 &  &  & Src. & E\_R & E\_I & E\_R & E\_I & E\_R & E\_I & E\_R & E\_I & EMOV \\
 \cline{5-13} 
 &  &  & Tgt. & E & E & E & E & LJ & LJ & V & V & E \\ \hline
Model & Tokenizer & Data & Cond. & \multicolumn{9}{c}{FFE} \\ \hline
\multicolumn{1}{c|}{\multirow{4}{*}{\begin{tabular}[c]{@{}c@{}}HuBERT\\(LS960)\end{tabular}}} & \multicolumn{1}{c|}{\multirow{2}{*}{\begin{tabular}[c]{@{}c@{}}KM500\\(LS960)\end{tabular}}} & \begin{tabular}[c]{@{}c@{}}E+LJ+V\end{tabular} & S & \begin{tabular}[c]{@{}c@{}}0.31$\pm$ 0.10\end{tabular} & \begin{tabular}[c]{@{}c@{}}0.33$\pm$ 0.12\end{tabular} & \begin{tabular}[c]{@{}c@{}}0.37$\pm$ 0.12\end{tabular} & \begin{tabular}[c]{@{}c@{}}0.38$\pm$ 0.13\end{tabular} & \begin{tabular}[c]{@{}c@{}}0.34$\pm$ 0.10\end{tabular} & \begin{tabular}[c]{@{}c@{}}0.35$\pm$ 0.12\end{tabular} & \begin{tabular}[c]{@{}c@{}}0.38$\pm$ 0.11\end{tabular} & \begin{tabular}[c]{@{}c@{}}0.38$\pm$ 0.14\end{tabular} & \begin{tabular}[c]{@{}c@{}}0.27$\pm$ 0.08\end{tabular} \\
\multicolumn{1}{c|}{} & \multicolumn{1}{c|}{} & \begin{tabular}[c]{@{}c@{}}E+LJ+V\end{tabular} & \begin{tabular}[c]{@{}c@{}} S\_E \end{tabular} & \begin{tabular}[c]{@{}c@{}}0.27$\pm$ 0.13\end{tabular} & \begin{tabular}[c]{@{}c@{}}0.30$\pm$ 0.13\end{tabular} & \begin{tabular}[c]{@{}c@{}}0.34$\pm$ 0.16\end{tabular} & \begin{tabular}[c]{@{}c@{}}0.36$\pm$ 0.15\end{tabular} & \begin{tabular}[c]{@{}c@{}}0.33$\pm$ 0.15\end{tabular} & \begin{tabular}[c]{@{}c@{}}0.35$\pm$ 0.14\end{tabular} & \begin{tabular}[c]{@{}c@{}}0.34$\pm$ 0.16\end{tabular} & \begin{tabular}[c]{@{}c@{}}0.36$\pm$ 0.15\end{tabular} & \begin{tabular}[c]{@{}c@{}}\textbf{0.25$\pm$ 0.09}\end{tabular} \\ \cline{2-13} 
\multicolumn{1}{c|}{} & \multicolumn{1}{c|}{\multirow{2}{*}{\begin{tabular}[c]{@{}c@{}}KM2000 (E)\end{tabular}}} & \begin{tabular}[c]{@{}c@{}}E+LJ+V\end{tabular} & S & \begin{tabular}[c]{@{}c@{}}0.31$\pm$ 0.12\end{tabular} & \begin{tabular}[c]{@{}c@{}}0.33$\pm$ 0.13\end{tabular} & \begin{tabular}[c]{@{}c@{}}0.36$\pm$ 0.13\end{tabular} & \begin{tabular}[c]{@{}c@{}}0.36$\pm$ 0.13\end{tabular} & \begin{tabular}[c]{@{}c@{}}0.35$\pm$ 0.09\end{tabular} & \begin{tabular}[c]{@{}c@{}}0.35$\pm$ 0.11\end{tabular} & \begin{tabular}[c]{@{}c@{}}0.38$\pm$ 0.11\end{tabular} & \begin{tabular}[c]{@{}c@{}}0.38$\pm$ 0.14\end{tabular} & \begin{tabular}[c]{@{}c@{}}0.26$\pm$ 0.09\end{tabular} \\
\multicolumn{1}{c|}{} & \multicolumn{1}{c|}{} & \begin{tabular}[c]{@{}c@{}}E+LJ+V\end{tabular} & \begin{tabular}[c]{@{}c@{}}S\_E\end{tabular} & \begin{tabular}[c]{@{}c@{}}\textbf{0.26$\pm$ 0.13}\end{tabular} & \begin{tabular}[c]{@{}c@{}}\textbf{0.29$\pm$ 0.13}\end{tabular} & \begin{tabular}[c]{@{}c@{}}0.34$\pm$ 0.16\end{tabular} & \begin{tabular}[c]{@{}c@{}}0.36$\pm$ 0.15\end{tabular} & \begin{tabular}[c]{@{}c@{}}\textbf{0.31$\pm$ 0.14}\end{tabular} & \begin{tabular}[c]{@{}c@{}}\textbf{0.33$\pm$ 0.13}\end{tabular} & \begin{tabular}[c]{@{}c@{}}\textbf{0.33$\pm$ 0.16}\end{tabular} & \begin{tabular}[c]{@{}c@{}}\textbf{0.35$\pm$ 0.15}\end{tabular} & \begin{tabular}[c]{@{}c@{}}0.26$\pm$ 0.09\end{tabular} \\
\hline
\multicolumn{1}{c|}{\multirow{6}{*}{\begin{tabular}[c]{@{}c@{}}HuBERT\\(Mix1)\end{tabular}}} & \multicolumn{1}{c|}{\multirow{3}{*}{\begin{tabular}[c]{@{}c@{}}KM2000(Mix1)\end{tabular}}} & \begin{tabular}[c]{@{}c@{}}E+LJ+V\end{tabular} & S & \begin{tabular}[c]{@{}c@{}}0.32$\pm$ 0.11\end{tabular} & \begin{tabular}[c]{@{}c@{}}0.33$\pm$ 0.13\end{tabular} & \begin{tabular}[c]{@{}c@{}}0.38$\pm$ 0.12\end{tabular} & \begin{tabular}[c]{@{}c@{}}0.37$\pm$ 0.13\end{tabular} & \begin{tabular}[c]{@{}c@{}}0.36$\pm$ 0.09\end{tabular} & \begin{tabular}[c]{@{}c@{}}0.35$\pm$ 0.11\end{tabular} & \begin{tabular}[c]{@{}c@{}}0.38$\pm$ 0.12\end{tabular} & \begin{tabular}[c]{@{}c@{}}0.38$\pm$ 0.14\end{tabular} & \begin{tabular}[c]{@{}c@{}}0.26$\pm$ 0.09\end{tabular} \\ 
\multicolumn{1}{c|}{} & \multicolumn{1}{c|}{} & \begin{tabular}[c]{@{}c@{}}E+LJ+V\end{tabular} & \begin{tabular}[c]{@{}c@{}}S\_E\end{tabular} & \begin{tabular}[c]{@{}c@{}}0.28$\pm$ 0.14\end{tabular} & \begin{tabular}[c]{@{}c@{}}0.29$\pm$ 0.13\end{tabular} & \begin{tabular}[c]{@{}c@{}}0.34$\pm$ 0.16\end{tabular} & \begin{tabular}[c]{@{}c@{}}0.36$\pm$ 0.15\end{tabular} & \begin{tabular}[c]{@{}c@{}}0.32$\pm$ 0.15\end{tabular} & \begin{tabular}[c]{@{}c@{}}0.34$\pm$ 0.14\end{tabular} & \begin{tabular}[c]{@{}c@{}}0.34$\pm$ 0.16\end{tabular} & \begin{tabular}[c]{@{}c@{}}0.36$\pm$ 0.15\end{tabular} & \begin{tabular}[c]{@{}c@{}}0.27$\pm$ 0.09\end{tabular} \\ \cline{2-13} 
\multicolumn{1}{c|}{} & \multicolumn{1}{c|}{\multirow{3}{*}{\begin{tabular}[c]{@{}c@{}}KM2000 (E) \end{tabular}}} & \begin{tabular}[c]{@{}c@{}}E+LJ+V\end{tabular} & S & \begin{tabular}[c]{@{}c@{}}0.31$\pm$ 0.10\end{tabular} & \begin{tabular}[c]{@{}c@{}}0.32$\pm$ 0.12\end{tabular} & \begin{tabular}[c]{@{}c@{}}0.37$\pm$ 0.12\end{tabular} & \begin{tabular}[c]{@{}c@{}}0.37$\pm$ 0.13\end{tabular} & \begin{tabular}[c]{@{}c@{}}0.35$\pm$ 0.09\end{tabular} & \begin{tabular}[c]{@{}c@{}}0.35$\pm$ 0.11\end{tabular} & \begin{tabular}[c]{@{}c@{}}0.38$\pm$ 0.11\end{tabular} & \begin{tabular}[c]{@{}c@{}}0.37$\pm$ 0.14\end{tabular} & \begin{tabular}[c]{@{}c@{}}0.26$\pm$ 0.08\end{tabular} \\
\multicolumn{1}{c|}{} & \multicolumn{1}{c|}{} & \begin{tabular}[c]{@{}c@{}}E+LJ+V\end{tabular} & \begin{tabular}[c]{@{}c@{}}S\_E\end{tabular} & \begin{tabular}[c]{@{}c@{}}0.27$\pm$ 0.13\end{tabular} & \begin{tabular}[c]{@{}c@{}}0.30$\pm$ 0.13\end{tabular} & \begin{tabular}[c]{@{}c@{}}\textbf{0.34$\pm$ 0.16}\end{tabular} & \begin{tabular}[c]{@{}c@{}}\textbf{0.36$\pm$ 0.14}\end{tabular} & \begin{tabular}[c]{@{}c@{}}0.32$\pm$ 0.14\end{tabular} & \begin{tabular}[c]{@{}c@{}}0.34$\pm$ 0.14\end{tabular} & \begin{tabular}[c]{@{}c@{}}0.34$\pm$ 0.16\end{tabular} & \begin{tabular}[c]{@{}c@{}}0.36$\pm$ 0.15\end{tabular} & \begin{tabular}[c]{@{}c@{}}0.25$\pm$ 0.09\end{tabular} \\ \hline
\multicolumn{2}{l|}{Encodec (RVQ-1)} & \_ & None & \begin{tabular}[c]{@{}c@{}}0.08$\pm$ 0.04\end{tabular} & \begin{tabular}[c]{@{}c@{}}0.11$\pm$ 0.07\end{tabular} & \_ & \_ & \_ & \_ & \_ & \_ & \begin{tabular}[c]{@{}c@{}}0.09$\pm$ 0.06\end{tabular} \\
\multicolumn{2}{l|}{Encodec (RVQ-8)} & \_ & None & \begin{tabular}[c]{@{}c@{}}\textbf{0.04$\pm$ 0.02}\end{tabular} & \begin{tabular}[c]{@{}c@{}}\textbf{0.05$\pm$ 0.03}\end{tabular} & \_ & \_ & \_ & \_ & \_ & \_ & \begin{tabular}[c]{@{}c@{}}\textbf{0.04$\pm$ 0.02}\end{tabular}\\
\bottomrule
\end{tabular}}
\vspace{-.7em}
\end{table*}

\vspace{-0.2cm}
\section{Results}
\vspace{-0.2cm}

Table \ref{tab:units} shows the phonetic quality metrics across different SSL units. The HuBERT encoders trained on the larger and noisier corpus (Mix1) tend to have overall better results than when trained on LS960 only, especially when tested on Fisher. The ABX-centroid and PNMI metrics gave better results when k-means clustering was run on \expresso~(a small high quality, high diversity dataset) than on the large dataset used to train HuBERT itself. This was not the case, however with the ABX 1-hot metric, so further study is necessary to confirm this result. The Encodec units gave poor results, which is not surprising given that Encodec units are generic representations trained for audio compression that also encodes non-phonetic variations whereas HuBERT units are trained with a masked language modeling objective.

Table~\ref{tab:wers} shows the result of content preservation for the resynthesis task (WER), distinguishing the case where the input sentences are drawn from the same distribution as the vocoder training sets (In-domain Source: LJ, VCTK and Exp-R) and when the input sentences are from different datasets (Out-of-domain Source: LS and Fisher). For in-domain, we further distinguish the cases where the input voice is the same as the target voice (Same Speaker) and when the target voice is randomly sampled from the same training set (Swapped Speaker; not applicable to LJ speech). We find that swapping speakers costs on average only a small decrement in WER (3\% relative), suggesting that the units are well (although not totally) disentangled from speaker information. On average, the HuBERT units trained on Mix1 give better performances (about 10\% relative) than units trained on LS960, a result consistent with the phonetic quality metrics. As for the vocoder, the training voices and conditioning (either speaker alone or speaker+style) did not give systematic results. Overall, the best HuBERT results on the same speaker showed a drop in performance compared to the original audio files (between 30\% relative to more than double the error), and compared to the Encodec-8 units. Regarding out-of-domain resynthesis, the HuBERT-Mix1 units again generally outperform the HuBERT-LS960 units (19\% relative). In addition, the tokenizer trained on \expresso~tend to be better by a small margin (3\% relative). The best resynthesis models suffer from a large drop in WER compared to original audio and Encodec-8 for LS (twice the errors) but a much smaller drop for Fisher.  Encodec-1 consistently underperform HuBERT-based resynthesis, indicating that one Encodec codebook of size 1024 is not enough to fully capture linguistic content.

Tables \ref{tab:EMO} and \ref{tab:FFE} show the results on style and pitch preservation, respectively; these experiments are exclusively ran on \expresso~inputs (in-domain) or EmoV inputs (out-of-domain). 
We group the in-domain results in 3 conditions. The first two are similar to the same-speaker and swapped speaker conditions discussed above. Unsurprisingly, for both style and pitch, the results are uniformly better when the vocoder was conditioned on speaker+expression than in speaker alone. Note, though, that even without expression conditioning, the style classification score is much higher than the chance level (3.8\%), suggesting that style is partly transmitted through the units. The results on swapped speakers are slightly worse than on same speaker (cost around 10\% relative accuracy on style and 20\% rel. on pitch error). Globally, the style scores of Encodec-8 are on par or better than the best HuBERT resynthesis models, but much better for pitch preservation (6-fold). Next, we explored whether expressivity could be transferred to voices that were only trained in the default read speech style (VCTK and LJ). This change cuts the style score in half, still remaining way above chance, and better than the \expresso~voices not conditioned by expression, showing that expressive styles can, to a certain extent, generalize to untrained voices in a zero-shot fashion. 
Finally, we tested expressive resynthesis could be applied out-of-domain, using input data from EmoV. This dataset uses 5 expressions (neutral, amused, angry, sleepy, disgust) which we mapped based on the description to \expresso's neutral, laughing, angry, sleepy and disgusted. The performance of the classifier was much lower than in the in-domain case, and unexpectedly higher for resynthesis than the original file. Inspection of the errors pattern showed some systematic style confusions across datasets. For instance, original EmoV voices in angry style were classified as ``projected'' by our classifier, but as ``angry'' once resynthetized with the ``angry'' conditioning. This suggests discrepancies in style rendering across the two datasets for identical labels (e.g., anger rendered as shouting in one versus cold rage in the other). This also suggests that the style-conditioned vocoder can to a certain extent remap input styles to different ones (confirmed by a style swapping experiment not reported in the table). More research and expressive datasets are needed to develop a dataset-independent and speaker-independent expressive style classifiers. Despite these limitations, the results were congruent with in the in-domain case, with better than chance performance and improved performance with expression-conditioned vocoders.

\vspace{-0.4cm}
\section{Conclusion}
\vspace{-0.2cm}
We presented a new dataset for expressive discrete resynthesis, and analysed content and style preservation for several baseline discrete SSL models. 
We showed that Encodec systems which are designed for general audio compression are generally better for resynthesis, although they lack the controllability in output voices and style made possible by the fact that HuBERT units are disentangled from speaker identity and (partially) from expressivity. Further work is needed to improve HuBERT-based expressive resynthesis, and reach the quality of Encodec units, while retaining controllability. In particular, improved performance on pitch preservation could be obtained by conditioning the vocoder on discrete pitch units, as in \cite{polyak21resynt}. 



\bibliographystyle{IEEEtran}

\bibliography{main-expresso.bib}

\end{document}